# GSLAM: Initialization-robust Monocular Visual SLAM via Global Structure-from-Motion

Chengzhou Tang
Simon Fraser University
cta73@sfu.ca

Oliver Wang
Adobe Research
owang@adobe.com

Ping Tan
Simon Fraser University
pingtan@sfu.ca

## Abstract

*Many monocular visual SLAM algorithms are derived from incremental structure-from-motion (SfM) methods. This work proposes a novel monocular SLAM method which integrates recent advances made in* global *SfM. In particular, we present two main contributions to visual SLAM. First, we solve the visual odometry problem by a novel rank-1 matrix factorization technique which is more robust to the errors in map initialization. Second, we adopt a recent global SfM method for the pose-graph optimization, which leads to a multi-stage linear formulation and enables L1 optimization for better robustness to false loops. The combination of these two approaches generates more robust reconstruction and is significantly faster (4×) than recent state-of-the-art SLAM systems. We also present a new dataset recorded with ground truth camera motion in a Vicon motion capture room, and compare our method to prior systems on it and established benchmark datasets.*

## 1. Introduction

Monocular visual Simultaneous Localization and Mapping (SLAM) [23, 16, 32] jointly solves for both the camera motion and 3D points (map) of an unknown environment from a single input video. Most recent methods take an optimization based approach, which typically consists of a front-end that uses visual odometry to track the camera, and a back-end that optimizes the map in a pose-graph. In real world use cases, both ends require careful initialization.

Visual odometry methods often resemble incremental structure-from-motion (SfM) [39] which initializes a 3D map from two frames and tracks the camera motion in realtime by image-to-map registration by PnP[31]. The initial 3D map is usually computed from a homography [17] or an essential matrix [33]. However, the model selection between homography and essential matrix is tricky, as discussed in [47, 32]. To make matters worse, the followed PnP based camera tracking heavily relies on the initial map, and visual SLAM systems cannot effectively start until find a suitable initial reconstruction. This problem crucially also makes it difficult to recover from tracking failures, when the system fails at a place unsuitable for map re-initialization.

The pose-graph component detects and optimizes loops to build a globally consistent map that does not suffer from drift. It is often optimized using a generic nonlinear least square solver, e.g. G2O [27] or Ceres [1]. One problem is that loop detection often generates false positives, which leads to outlier camera motion constraints. Many robust optimization schemes [8, 30, 43, 28] have therefore been proposed to deal with false loop closures by introducing switching variables or different weighting schemes, but their effectiveness heavily relies on the initialization of the switching variables or weights.

In summary, visual odometry and pose-graph optimization are fundamentally sensitive to initialization, as they iteratively optimize a highly non-convex objective function [4], where only local convergence is guaranteed. A similar challenge appears in the related problem of SfM, where the final bundle adjustment (BA) [49] iteratively minimizes the non-convex re-projection error of all points, also requiring careful initialization (e.g. solving the next-best-view problem [21]). This initialization problem has been intensively studied in recent global SfM methods [22, 53, 36, 12], and these methods have been proven to deliver state-of-the-art accuracy and efficiency in SfM.

To that end, we propose a novel visual odometry method inspired by recent global SfM techniques. Since global SfM methods are designed to operate on pre-captured images, we propose an odometry modification that works on real-time input videos within an expanding sliding window. Most importantly, we introduce a novel rank-1 matrix factorization that is able to compute the camera positions and scene points together, given a rough camera rotations. The camera rotations can be computed separately, for example by the methods in [26, 25], however, the robustness of our approach to noisy rotations allows us to also use noisy inertial measurement unit (IMU) readings when available, greatly improving running time. We empirically demonstrate that our factorization based method is more robust than the traditional PnP based camera tracking, and it is ad-ditionally more robust to noisy correspondences. Furthermore, using our approach, each incoming frame will trigger the factorization, thus outliers in earlier frames are not fixed, and can be corrected if later frames bring in more inliers.

We also present a novel pose-graph following recent global SfM methods [22, 53, 36, 12]. We solve the camera rotation, scale and translation separately. Separating the rotation and translation estimation brings advantages to 2D SLAM problems [5, 7, 6]. We generalize that to the 3D case here, leveraging a recent multi-stage linear method [12] to solve the pose-graph optimization. Each stage, has a closed-form solution and does not reply on initialization. The linear formulation also allows us to adopt robust L1 optimization techniques [50], which helps to deal with false loops. Furthermore, to speed up computation for real-time SLAM, we solve the L1 optimization using an incremental strategy.

In summary, this paper exploits recent global SfM techniques to make visual SLAM more robust to initialization problems. To demonstrate the effectiveness of our method, we introduce a dataset of 10 video sequences with ground truth camera poses captured in a motion capture room, which we plan to release publicly. We show that our method performs favorably when compared to state-of-the-art methods, including finely tuned systems such as ORB-SLAM[32] and LSD-SLAM[16], on both our new ground truth dataset as well as other public benchmark dataset such as TUM-RGBD [41], at significantly less (approximately $4\times$ less) running time. We will release the source code for others to evaluate on their data as well.

## 2. Related Work

Many visual SLAM algorithms have been proposed in recent years, and a comprehensive review is beyond the scope of this work. We discuss some most relevant and recent works here, and refer the audience to a recent survey [4] for a more thorough discussion.

**Map initialization in odometry.** Early visual SLAM algorithms [14, 15, 11, 10] are based on extended Kalman filters, where 3D map points are initialized with large uncertainty and then update the 3D map when a new frame is inputed. Keyframe-based methods (also known as optimization-based methods) [34, 37, 23, 32] solve the SLAM problem in an incremental SfM framework. Some of the odometry methods [23, 18] assume that cameras start observing a planar scene, in order to initialize the SfM from a homography. Others like [44] initialize from an essential matrix to deal with more general scenes. However, the essential matrix estimation [33] itself generates multiple solutions and requires careful solution selection. Furthermore, during initialization, the selection between homography and essential matrix is also tricky, as discussed in the ORB-SLAM system [32]. Therefore, ORB-SLAM evaluates a model confidence and is only able to start reconstruction once the uncertainty is sufficiently small.

The recent direct method LSD-SLAM [16] randomly initializes the 3D map. In this way, LSD-SLAM is free from initialization selection problems. However, its accuracy is often inferior as reported in [32] and as observed in our experiments when it diverged from the initialization.

As opposed to the above approaches, we address the map initialization problem by exploiting global SfM methods for visual odometry. Our method is free from model selection and optimizes all involved cameras and points simultaneously. As shown in experiments, this allows us to robustly initialize SLAM in challenging situations.

**Map optimization via the pose-graph.** The map optimization is often solved by a nonlinear least square problem, using advanced libraries such as G2O [27] or Ceres [1], which are designed to efficiently minimize the linearized objective function through an iterative approach. Since the objective function is non-convex, a good initialization is critical to the nonlinear optimization in pose-graph. Direct initialization by visual odometry alone usually produces poor results due to drifting errors, so loop closures [52] are used to reduce the drifting. Usually, loops are detected by bag-of-words (BoW) based method like [35, 13, 19]. However, these approaches still often generate false loops, which causes incorrect relative motion estimations and makes the pose-graph optimization fail. The effect of false loops can be reduced by introducing switch variables [43], internal consistency checks [28], or iteratively updated loop weights [30]. All these methods require a good initialization of the loop weights or switching variables.

Follow recent global SfM methods [12], we separate the computation of camera rotations and translations. In this way, our pose-graph optimization becomes a multi-stage linear problem, where at each stage we minimize the L1 norm of the linear objective function, which is more robust to false loops in the pose-graph. As shown in experiments, our method can deal with false loops that arise, for example, due to repetitive objects in the scene.

## 3. Overview

Our monocular SLAM system estimates the 3D map using a pose-graph of keyframes and estimates poses of other frames relative to these keyframes. As show in Figure 1, our system has three major components: **feature tracking**, **visual odometry** to track camera motion and reconstruct a local 3D map for each keyframe, and **pose-graph optimization** to estimate a global consistent map over all keyframes.

We use existing methods for feature tracking, detecting ORB [38] features on keyframes, which are good for matching features between keyframes from different viewpoints as shown in [32]. We then track ORB features across frames using the forward-additive KLT method [2].

For visual odometry, we proposed a novel fomulation that can be solved by rank-1 factorization in Section 4, It recieves images in an expanding sliding window and reconstructs a local 3D map associated with a keyframe in the window. For pose-graph optimization, we modified the offline method in [12] to an online version which stitches local maps from different keyframes together to form a globally consistent map in Section 5.

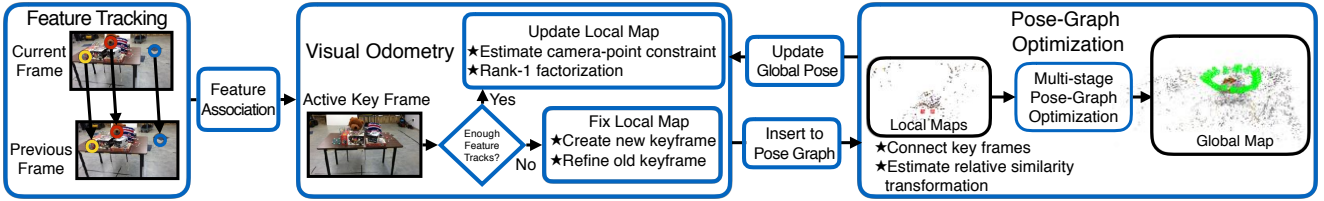

Figure 1. Overview of our monocular SLAM approach, which consists of **feature tracking** between neighboring frames, an novel rank-1 **visual odometry** solution for frames within an expanding local sliding window, and a global **pose-graph optimization** for all keyframes.

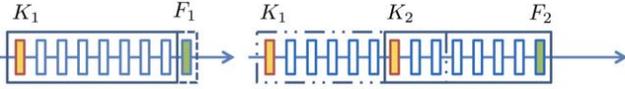

Figure 2. Left: when sufficient feature points are tracked to a new frame $c$, the local window attached to the keyframe $K_1$ expands to include $F_1$. Right: when insufficient features are tracked, a new keyframe $K_2$ will be selected, and a new local window expands to cover $F_2$.

## 4. Visual Odometry by Rank-1 Factorization

Our **visual odometry** reconstructs all frames within a local sliding window using a *linear* algorithm with guaranteed convergence, which is much more robust to map initialization than traditional PnP based method, making it appealing for monocular SLAM.

### 4.1. Local Sliding Window

When a new frame comes, we track the feature points from its previous frame. If more than 30% of features can be tracked, the local sliding window will expand to include the current new frame. Otherwise, the original local window is closed and its key-frame will be sent to the posegraph for subsequent optimization. At the same time, a new keyframe is created using the most recent frame that spans a sufficient baseline with the new frame. Here, the baseline is measured as the median angle between the view rays of corresponding feature points. The local window anchored to the new keyframe starts with it, and extends to the most recently seen frame. This process is illustrated in Figure 2.

### 4.2. Rank-1 Factorization

**Local Map Representation.** In the $i$-th local window, the local map start with a keyframe $K_i$ as the reference view and expands with a sequence of $m$ subsequent frames $F_{j=1\cdots m}$, and $n$ scene points $\mathbf{P}_{k=1\cdots n}$ visible in all the $m$ frames. In a local map anchored at keyframe $K_i$, $K_i$'s rotation $\mathbf{R}_i^{(i)}$ is considered to be identity and its camera position $\mathbf{c}_i^{(i)}$ is fixed at the origin. The superscript index $^{(i)}$ indicates quantities in the local map of $K_i$. The camera poses of other frames $F_{j=1\cdots m}$ in the local map are defined by rotations $\mathbf{R}_{j=1\cdots m}^{(i)}$ and camera positions $\mathbf{c}_{j=1\cdots m}^{(i)}$. We use inverse depth to parameterize scene points in the local map of $K_i$ and have $\mathbf{P}_k^{(i)} = \frac{1}{d_k^{(i)}} \mathbf{p}_k^{(i)}$, where $\mathbf{p}_k^{(i)} = \frac{\mathbf{P}_k^{(i)}}{\|\mathbf{P}_k^{(i)}\|}$ is the view ray of $\mathbf{P}_k^{(i)}$ on $K_i$ and is computed by normalizing the corresponding feature point using intrisic matrix, $d_k^{(i)}$ is the unkown inverse depth. The ray of $\mathbf{P}_k^{(i)}$ on $F_j$ is $\mathbf{p}_k^{(j)}$.

**Relative Motion Between Frames.** We present a novel global SfM method to estimate all camera poses and 3D scene points within a local window. Our global SfM method takes the relative motions between frames as input. First, we determine the relative camera rotation $\mathbf{R}_j^{(i)}$ between keyframe $K_i$ and frame $F_j$ by [26] or integrating IMU readings when available. Given rotations and a set of corresponding features, we compute the translation $\mathbf{t}_j^{(i)}$ between $K_i$ and $F_j$ up to a scale by two-point algorithm [29].

**Camera-Point Constraint.** Next, we derive a constraint for frame $F_j$'s camera position $\mathbf{c}_j^{(i)}$ and a point $\mathbf{P}_k^{(i)}$. We first assume $\mathbf{P}_k^{(i)}$ has unit depth, i.e. $\mathbf{P}_k^{(i)} = \mathbf{p}_k^{(i)}$ to make the derivation simple. This will be easily generalized later. We compute $\mathbf{p}_{jk}^{(i)}$ in Figure 3, which is the direction from $\mathbf{c}_j^{(i)}$ to $\mathbf{p}_k^{(i)}$, as:

$$\mathbf{p}_{jk}^{(i)} = \mathbf{R}_j^{(i)\top} \mathbf{p}_k^{(j)}. \qquad (1)$$

Ideally, without noise in everything, we have

$$\mathbf{c}_j^{(i)} = \alpha_{jk} \mathbf{t}_j^{(i)} = \mathbf{p}_k^{(i)} - \beta_{jk} \mathbf{p}_{jk}^{(i)}. \qquad (2)$$

Equation (2) only has two unknowns $\alpha_{jk}$ and $\beta_{jk}$ which are the distance between $\mathbf{c}_j^{(i)}$ and $\mathbf{c}_i^{(i)}$, and the distance between $\mathbf{c}_j^{(i)}$ and $\mathbf{p}_k^{(i)}$ respectively as shown in Figure 3. However, Equation (2) rarely holds because noise exists. Instead, $\mathbf{c}_j^{(i)}$ can be approximated as the mid-point of the line segment **ab** as shown in Figure 3, which is the mutal perpendicular line segment to the two rays $\mathbf{t}_j^{(i)}$ and $-\mathbf{p}_{jk}^{(i)}$:

$$\mathbf{c}_j^{(i)} \approx \frac{1}{2}(\mathbf{a} + \mathbf{b}) = \frac{1}{2}(\alpha_{jk} \mathbf{t}_j^{(i)} + \mathbf{p}_k^{(i)} - \beta_{jk} \mathbf{p}_{jk}^{(i)}). \qquad (3)$$

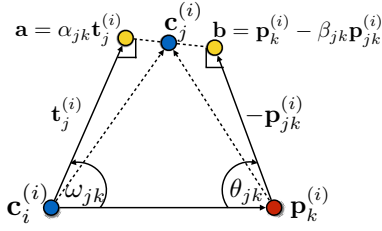

Figure 3. The geometric constraint of the position between one scene point $\mathbf{p}_k^{(i)}$ and two cameras $\mathbf{c}_i^{(i)}$ and $\mathbf{c}_j^{(i)}$. Please refer to the text for more details.

To estimate $\alpha_{jk}$ and $\beta_{jk}$, we solve

$$\begin{bmatrix} \mathbf{t}_j^{(i)\top} \\ -(\mathbf{p}_{jk}^{(i)})^\top \end{bmatrix} (\mathbf{a} - \mathbf{b}) = \mathbf{0}_{2\times 1} \qquad (4)$$

in closed-form because Equation (4) is a $2 \times 2$ linear system with 2 unknowns.

Now, all the variables are known in Equation (3), we can establish the relation between $\mathbf{c}_j^{(i)}$ and $\mathbf{p}_k^{(i)}$ by replacing $\mathbf{t}_j^{(i)}$ as $\mathbf{R}(\omega_{jk})\mathbf{p}_k^{(i)}$ and replacing $\mathbf{p}_{jk}^{(i)}$ as $\mathbf{R}(\theta_{jk})\mathbf{p}_k^{(i)}$:

$$\mathbf{c}_j^{(i)} \approx \frac{1}{2}\left(\alpha_{jk}\mathbf{R}(\omega_{jk})\mathbf{p}_k^{(i)} + \mathbf{p}_k^{(i)} - \beta_{jk}\mathbf{R}(\theta_{jk})\mathbf{p}_k^{(i)}\right), \quad (5)$$

where $\mathbf{R}(\omega_{jk})$ rotates $\mathbf{p}_k^{(i)}$ to $\mathbf{t}_j^{(i)}$ around the axis $\mathbf{t}_j^{(i)} \times \mathbf{p}_k^{(i)}$ and $\mathbf{R}(\theta_{jk})$ rotates $\mathbf{p}_k^{(i)}$ to $\mathbf{p}_{jk}^{(i)}$ around the axis $\mathbf{p}_{jk}^{(i)} \times \mathbf{p}_k^{(i)}$, which is known as the "rotation tricks" [22, 12]. Equation (5) can be further simplified as:

$$\mathbf{c}_j^{(i)} \approx \mathbf{A}_{jk}\mathbf{p}_k^{(i)}, \qquad (6)$$

where $\mathbf{A}_{jk} = \frac{1}{2}\left(\alpha_{jk}\mathbf{R}(\omega_{jk}) + \mathbf{I} - \beta_{jk}\mathbf{R}(\theta_{jk})\right)$ is a known $3 \times 3$ matrix that transforms the scene point's projection $\mathbf{p}_k^{(i)}$ to the camera position $\mathbf{c}_j^{(i)}$, and Equation (6) can be easily generalized to a point $\mathbf{P}_k^{(i)}$ with arbitary inverse depth $d_k^{(i)}$ as:

$$\mathbf{c}_j^{(i)} \approx \mathbf{A}_{jk}\mathbf{P}_k^{(i)} = \frac{1}{d_k^{(i)}}\mathbf{A}_{jk}\mathbf{p}_k^{(i)} = \frac{\mathbf{v}_{jk}}{d_k^{(i)}}, \qquad (7)$$

where $\mathbf{v}_{jk} = \mathbf{A}_{jk}\mathbf{p}_k^{(i)}$ is a known vector. Then we can solve for all cameras and all points' inverse depths by:

$$\underset{\substack{\mathbf{c}_{j=1\cdots m}^{(i)} \\ d_{k=1\cdots n}^{(i)}}}{\text{minimize}} \quad \sum_{j=1}^{m}\sum_{k=1}^{n} \|\mathbf{c}_j^{(i)}d_k^{(i)} - \mathbf{v}_{jk}\|_2^2, \qquad (8)$$

where $\|\mathbf{c}_j^{(i)}d_k^{(i)} - \mathbf{v}_{jk}\|_2^2$ is a reweighted geometric error that gives the closer points higher confidence by multiplying $d_k^{(i)}$ to the camera position $\mathbf{c}_j^{(i)}$ and $\frac{\mathbf{v}_{jk}}{d_k^{(i)}}$.

**Rank-One Factorization.** To solve Equation (8), we derive a factorization-based method which reconstructs *all* cameras and *all* scene points simultaneously in a local window. Specifically, we assemble all the vectors $\mathbf{v}_j^k$ into a matrix $\mathbf{M}$ and then factorize $\mathbf{M}$ to estimate the camera positions and scene points' depths together:

$$\begin{aligned}
\mathbf{M}_{3m\times n} &= \begin{bmatrix} \mathbf{v}_{1,1} & \cdots & \mathbf{v}_{1,n} \\ & \vdots & \\ \mathbf{v}_{m,1} & \cdots & \mathbf{v}_{m,n} \end{bmatrix} = \begin{bmatrix} \mathbf{c}_1^{(i)}d_1^{(i)} & \cdots & \mathbf{c}_1^{(i)}d_n^{(i)} \\ & \vdots & \\ \mathbf{c}_m^{(i)}d_1^{(i)} & \cdots & \mathbf{c}_m^{(i)}d_n^{(i)} \end{bmatrix} \\
&= \begin{bmatrix} \mathbf{c}_1^{(i)} \\ \mathbf{c}_2^{(i)} \\ \vdots \\ \mathbf{c}_m^{(i)} \end{bmatrix} \begin{bmatrix} d_1^{(i)} & d_2^{(i)} & \cdots d_n^{(i)} \end{bmatrix} = \mathbf{C}_{3m\times 1}\mathbf{D}_{1\times n},
\end{aligned}$$
(9)

where $\mathbf{M}_{3m\times n}$ is a rank-1 matrix and factorized as the product of the cameras $\mathbf{C}_{3m\times 1}$ and the inverse depths $\mathbf{D}_{1\times n}$. We factorize Equation (9) by SVD. Since our factorization is rank-1, the solution is dertermined only up to a scale factor.

We note that our factorization works for *perspective* cameras, while the classic factorization methods only deal with orthography or affine cameras [46, 42] or require knowing the 'projective depth' for perspective cameras [48]. Furthermore, our rank-1 factorization can work with as few as two frames, which makes it robust to feature occlusion. In summary, our rank-1 factorization works as:

---
**Algorithm 1:** Odometry by Rank-1 Factorization at $F_j$

---
**Input:** $\mathbf{p}_{k=1\cdots n}^{(i)}$, $\mathbf{p}_{k=1\cdots n}^{(j)}$, and $\mathbf{M}$ computed at $F_{j-1}$
**Output:** relative rotation $\mathbf{R}_j^{(i)}$, relative positions $\mathbf{c}_{j=1\cdots m}^{(i)}$ and scene points' inverse depths $d_{k=1\cdots n}^{(i)}$
1: Estimate rotation $\mathbf{R}_j^{(i)}$ by [26] or IMU integration.
2: Esimate translation $\mathbf{t}_j^{(i)}$ by two-point algorithm [29].
3: Rotate $\mathbf{p}_{k=1\cdots n}^{(j)}$ by $\mathbf{R}_j^{(i)\top}$ as in Equation (1).
4: **for** $k = 1$ to $n$ **do**
5:    Esimate $\alpha_{jk}$ and $\beta_{jk}$ by solving Equation (4)
6:    Compute $\mathbf{v}_{jk}$ by Equation (5) to (7)
7: **end for**
8: Update $\mathbf{M}$ by adding $[\mathbf{v}_{j1}, \mathbf{v}_{j2}, \cdots, \mathbf{v}_{jm}]$ as new rows and factorize it as in Equation (9).

---

**Local Map Refinement.** Our factorization only reconstructs features that are tracked through *all* frames in the local window. For better SLAM performance, it can be preferable to reconstruct more feature points in each local window. Therefore, when the number of factorized 3D points is less than 30% of all the feature points on the keyframe, we triangulate the remaining feature tracks that only partially cover the local window. Finally, we refine all points in the local map with BA. We only do the local BA when a keyframe is sent to the pose-graph for efficiency.

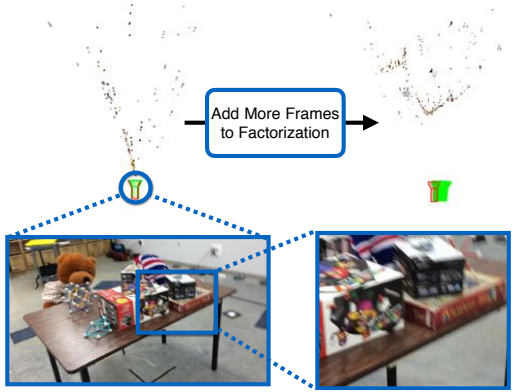

Figure 4. Our factorization is robust to initialization error. Left: the initial local 3D map is corrupted because one of the the input frames is blurry and has poor feature correspondences. Right: after including more clear frames, the local 3D map converges to a better shape because the proposed rank-1 factorization has guaranteed convergence.

## 4.3. Discussion and Evaluation

**Convergence** The rank-1 factorization defined in Equation (9) can be solved as a SVD problem by power method [20]. There exists an constant $\alpha$ such that

$$\|\mathbf{M} - \mathbf{C}_k \mathbf{D}_k\| \leq \alpha (r_2/r_1)^{2k}, \qquad (10)$$

where $\mathbf{C}_k$ and $\mathbf{D}_k$ are the solutions after $k$ iterations. If the matrix $\mathbf{M}$ is exactly rank-1 the algorithm would converge in one iteration because the second largest singular value $r_2 = 0$. In practice, our rank-1 factorization converges in 2-3 iterations because $r_2$ is usually $100\times$ smaller than the largest singular value $r_1$. So the convergence of our rank-1 factorization is both gauranteed and fast. In our experiments, it takes 4ms per frame. Besides the efficiency, it is appealing for visual SLAM because of the initialization robustness.

**Initialization-Robustness** As we discussed in Section 1, visual odometry based on incremental SfM algorithms is sensitive to map initialization while our global SfM based algorithm is robust to initialization. For example, as illustrated in Figure 4, an earlier frame in the local window is blurry and has noisy feature tracks, degrading the relative pose and local map initialization. However, because of the gauranteed convergence of our rank-1 factorization, the map converges to a reasonable solution after adding more frames, as shown on Figure 4, right. Similar problem also happens when IMU is unavailable. The rotation estimated by [26] is sometimes unstable for the first several frames. However, our factorization is robust to the errors at the begining and can converge to a reasonbale results as long as more correct estimations are inserted to a window.

**Evaluation on Synthetic Sequences** To further validate the advantage of initialization-robustness, we conducted experiments on synthetic data with known ground truth. We follow the experiment design in [33, 22]. To simulate the condition of initialization in visual SLAM, we genereate synthetic video clips of 30 frames. The first camera is at the origin with identity rotation as the keyframe's pose. The scene points are generated within the view frustum of the first camera with a minimum depth of 5 unit and maximum depth of 10 units as the close setting, or with minimum depth of 10 units and maximum depth of 15 units as the far setting. The camera trajectory is generated as either circular motion or forward motion towards the center of the points, and the invterval between neighbouring cameras was 0.05 units. The horizontal FoV of the cameras are $60°$, and the image resolution is $800\times600$ pixels. We perturb the image coordinates by zero-mean Gaussion noise with standard deviation in 3 pixels.

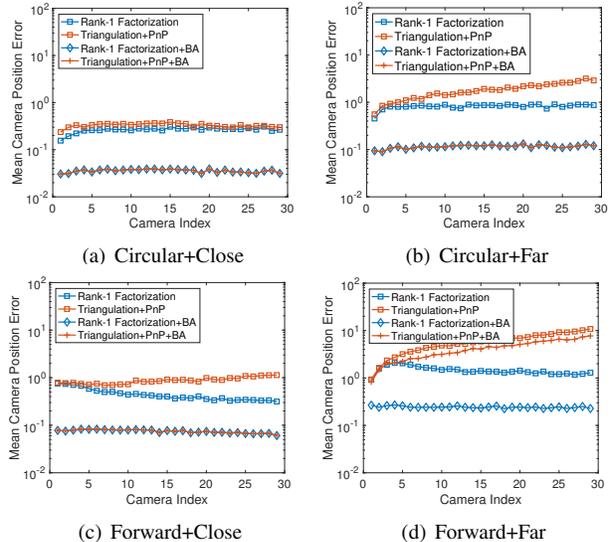

(a) Circular+Close  (b) Circular+Far

(c) Forward+Close  (d) Forward+Far

Figure 5. Syntheitc evaluation of initialization robustness on circular and forward camera motion with close and far points.

On the generated synthetic data, we compare the accracy of our rank-1 factorization and traditional visual SLAM based on incremental SfM, i.e. initialize the map from two views and use PnP to track the additional views. We measure the camera position error, normalized by the mean distance between neighboring cameras. For our method, we run our rank-1 factorization in an initialization-free way by staring immediately from the second camera. For Triangulation+PnP approach, we follow ORB-SLAM to initial map by triangulation when the median angle between the view rays of corresponding feature points is larger than $1.15°$ and estiamte the following cameras using EPnP [31]. Note there are algorithms (e.g. [45, 3]) to determine the optimal image pair for map initialization. But those methods are rarely used in visual SLAM because they often require access of the whole image sequence beforehand. Figure 5 shows the mean error of each camera after 1000 trials for each setting. Both methods achieve reasonbale results on close scene points, and converges to the same solutions after BA. On far scene points, our method is much more accurate. For circular motion, they still converge to the same solutions. But for forward motion, BA fails to converge to a correct solution with initialization from triangulation+PnP while successfully converges with initialziation from our rank-1 factorization. Foward motion is known as a difficult case for monocular SLAM. It induces a large number of local minima in the re-projection error [51], which makes BA highly dependent on the initialization. Therefore, visual SLAM algorithms often have trouble initializing with forward motion (very common in videos), especially when the parallax is small.

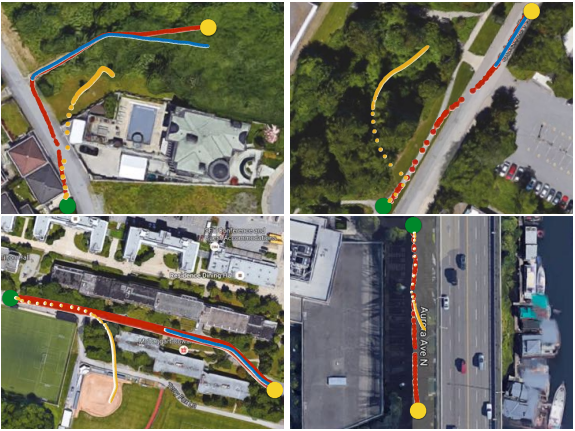

Figure 6. Map initialization comparison on real videos. Camera trajectories of our method (red dots), ORB-SLAM (blue dots), and LSD-SLAM (orange dots) are shown respectively. The big green and yellow dots are the ground truth starting and ending points, manually refined from GPS data.

In addition to being more accurate, our rank-1 factorization is very efficient. In our experiments, BA initialized with our rank-1 factorization converges in 2-3 iterations, making it very efficient, and indicating that our factorization produces results close to the optimal solution. Meanwhile, BA with triangulation+PnP converges in about 10-23 iterations because the initialization is far from optimal solution.

**Evaluation on Real Sequences** To show the difficulty of initialization on real sequences, we captured 4 walking sequences with forward motion using a mobile phone and compare our method with ORB-SLAM and LSD-SLAM. We run our approach (GlobalSLAM), ORB-SLAM, and LSD-SLAM 10 times on each of these sequences and average the results. Our system consistently performs well, while ORB-SLAM and LSD-SLAM tend to produce poor results. In Figure 6, we visualize typical results from different systems. Keyframe trajectories recovered by our method (red) are approximately aligned with the ground truth start (the green dot) and end (the yellow dot) locations of the videos, as shown on a satellite image. In comparison, ORB-SLAM trajectories (blue) only start at frames with sufficiently large parallax, only reconstructed 41.6%, 24.2% and 42.9% of the frames in the first 3 sequences and exhibit a total failure on the last sequence. In these experiments, we found ORB-SLAM tends to select a homography model to initialize due to the small parallax. It triangulates the scene points as in [24] and refines them by BA, however, the BA converges to a local minimum because the initialization is too far from the global minimum. ORB-SLAM will then detect this failure, reject the wrong map initialization and drop all the previous frames. In these experiments, ORB-SLAM repeats this procedure and refuses to start until the parallax is sufficiently large.

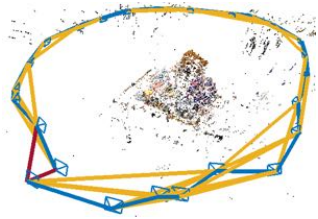

Figure 7. Pose-graph formed by keyframes. The edges of different colors indicate different kinds of keyframe connections, blue for direct neighbors, orange for extended neighbors, and red for loops. Please see text for more details.

On these four videos, the initialization of LSD-SLAM is even more unstable as it minimizes the photometric error (see similar discussion in [32]), which contains much more local minima than the geometric error. As a result, LSD-SLAM failed on these sequences and generated the trajectories shown by the orange dots in Figure 6.

## 5. Multi-Stage Linear Pose-Graph

Our **pose-graph optimization** computes a globally consistent map over all keyframes, using the local maps from the odometry. The optimization is triggered every time a new keyframe is inserted.

### 5.1. Pose-Graph Construction.

We start by describing how we build the pose graph. The pose-graph contains a set of vertices corresponding to keyframes, and edges, which indicate that the 3D similarity transformation between the pair of keyframes' local maps is known [40]. The similarity transformation between $K_i$ to $K_j$ consists of a scaling $s_{i,j}$, a rotation $\mathbf{R}_{i,j}$, and a relative camera position $\mathbf{c}_{i,j}$. Specifically, $\mathbf{R}_{i,j} = \mathbf{R}_j^{(i)}$ and $\mathbf{c}_{i,j} = \mathbf{c}_j^{(i)}$. To determine the scale change $s_{i,j}$, we collect feature correspondences between $K_i$ and $K_j$. Suppose $\mathbf{P}_k$ is a corresponding 3D point, we use its distance to the two cameras to estimate the relative scale change as:

$$s_{i,j} = \frac{\|\mathbf{P}_k^{(j)} - \mathbf{c}_i^{(j)}\|_2}{\|\mathbf{P}_k^{(i)} - \mathbf{c}_i^{(i)}\|_2}. \tag{11}$$

Here, $^{(i)}$ and $^{(j)}$ indicates quantities in the local map of $K_i$ and $K_j$. As each corresponding point generates a relative scale estimation, we take the median value for robustness. We connect keyframes according to the following three conditions.

First, we connect two neighboring keyframes $K_i$ and $K_{i+1}$. Typically, $K_{i+1}$ appears in the local window attached to $K_i$ as shown in Figure 2. Therefore, the relative motion between them is known. As shown in Figure 7, we use blue edges to link neighboring keyframes.

Second, we extend the neighborhood to increase the density of edges in the pose-graph. Specifically, if $K_j$ and $K_i$ are connected, we connect $K_j$ and $K_{i+1}$ if they have more than 50 matched features. As shown in Figure 7, extended neighbors are linked by orange edges. To determine the relative motion between $K_j$ and $K_{i+1}$, we use the pose of $K_j$ in the local map of $K_i$ (which is known since $K_j$ and $K_i$ are connected already). As $K_{i+1}$ is also in the local map of $K_i$, we can get the relative rotation and translation from $K_j$ to $K_{i+1}$ by concatenating the motion from $K_j$ to $K_i$ and $K_i$ to $K_{i+1}$. The relative scale $s_{j,i+1}$ is also computed from corresponding 3D points.

Finally, we also link keyframes by loop detection [19] (red edges in Figure 7). To determine the relative motion, we match feature points between the two keyframes and estimate their relative motion by EPnP [31]. This relative motion is then refined by BA for better robustness.

**Incremental Graph Optimization** We adopt the multi-stage linear method in [12] to solve our graph optimization problem. This method solves for a global rotation, scale, and translation separately, each as a linear problem. These linear sub-problems minimize geometric errors with L1-Norm in SO(3) for rotation [9] and in Euclidean space for scale and translation [12], all of which can be solved efficiently with linear programming. To fit realtime SLAM applications, we adopt an incremental approach in linear programming. When a new keyframe is inserted, we speed up the pose-graph optimization by initializing the linear programming with the solution at the previous keyframe. Please refer to the supplementary materials for details.

### 5.2. Evaluation on Real Sequences

By solving the problem using linear programming in multiple stages, our pose-graph is robust to false loop detections. A similar conclusion was observed in [12] on Internet images. To further show this advantage in SLAM, we compared our result to the SIM3 pose-graph with Huber loss in G2O [27, 40], which is adopted in both ORB-SLAM and LSD-SLAM. We also compare with the expectation maximization (EM) based robust pose-graph in [30]. To fairly evaluate the effect of our linear pose-graph optimization in isolation, we used the same components other than pose-graph optimization in all comparisons.

Figure 8 shows two scenes with repetitive structures, where the BoW based loop detection method generates false loops. To visualize the repetition, we generate a panorama of the scene and show two video frames that observe the same structures. As shown in Figure 8, our pose-graph is robust to false loops and produces a reasonable keyframe trajectory, as it minimizes the L1-norm, which is less prone to outliers. The false loops (red line segments) are detected and marked as dashed lines by our system. Both G2O and the EM method [30] on the other hand, generate incorrect results, failing to exclude the incorrectly detected loop. A possible reason is that these two methods estimate rotation, scale, and translation together, which hinders their ability to quantify motion residuals, generating wrong weights.

## 6. Quantitative Evaluation

In this section, we compare the quantitative results of our method to the widely used state-of-the-art systems ORB-SLAM [1] and LSD-SLAM on both the public TUM-RGBD bechmark [41] and our own dataset with ground truth camera poses recorded in a Vicon room. All the experiments are run on a 2013 MacBook Pro with 2.3GHz i7 CPU and 8GB memory. Our system runs at approximately 40 fps on 1080P videos with 3,000 tracked feature points per frame, which is roughly 4× faster than both ORB-SLAM and LSD-SLAM.

We captured 10 video sequences with ground-truth camera motion in a Vicon room. These sequences are captured in 1080p include different camera movement, such as circular motion, forward motion, and scenes with moving objects. All data will be made available to the public, along with the code to encourage follow-up research. Figure 9 shows some of the examples, with a sample frame above and the reconstructed 3D map below. Please see the supplemental material for videos and results presented here.

We compute the average keyframe position error for our method, ORB-SLAM, and LSD-SLAM. As LSD-SLAM starts from a random initialization, we skip its first 150 frames when computing the average keyframe position error. Generally speaking, as shown in Table 1, our method generates similar accuracy as ORB-SLAM and is more precise than LSD-SLAM in these examples. We run each method for 10 times on each sequence, and list the median accuracy and per frame time in Table 1.

As shown in Table 1, our approach performs comparable to or slightly better than ORB-SLAM, while being significantly (about 4×) faster than both ORB-SLAM and LSD-SLAM. On our system, our method takes on average, 20.6 ms to process a frame, while ORB-SLAM and LSD-SLAM take 85.4 and 76.4 ms respectively.

We also conduct quantitative comparisons on the TUM-RGBD benchmark dataset [41]. Note this dataset does not have IMU data. So we use the method in [26] to estimate the relative rotation between frames. The results of LSD-

---
[1]We use ORB-SLAM2, the enhanced version of ORB-SLAM, for all experiments.

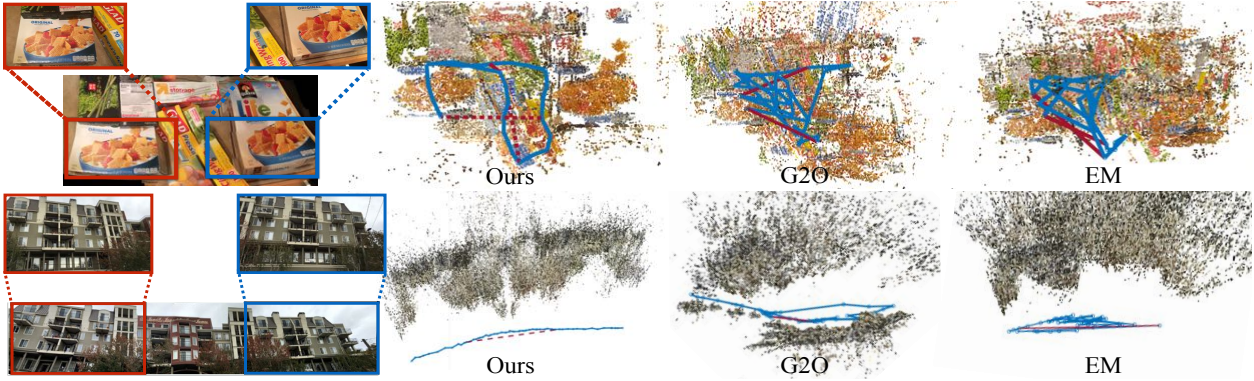

Figure 8. Comparison on videos with falsely detected loops. On the left are panoramas from two sample videos, along with two video frames. Right are the 3D reconstructions with keyframe trajectories optimized by our pose-graph, G2O[27, 40], and the EM method [30].

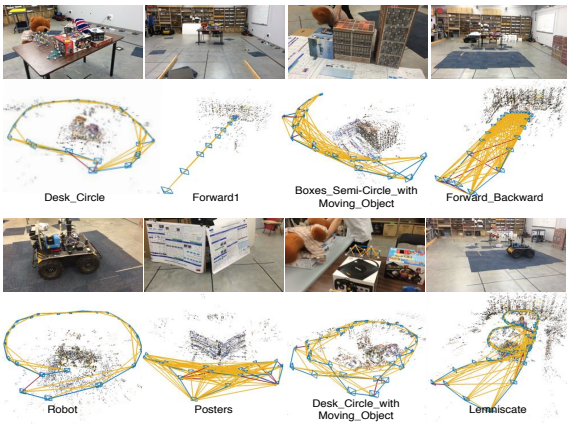

Figure 9. Results on some of the Vicon sequences with ground truth poses. For each example, we show an input frame and the recovered 3D global map.

| | | KF Trajectory RMSE (cm) | | |
|---|---|---|---|---|
| Sequence | | Ours | ORB | LSD |
| fr1xyz | | **0.6** | 0.9 | 9.0 |
| fr1desk | | 3.8 | **1.7** | 10.7 |
| fr2xyz | | **0.2** | 0.3 | 2.2 |
| fr2desk | | 2.1 | **0.9** | 4.6 |
| fr3household | | 3.5 | **2.2** | 38.0 |
| fr3sittingxyz | | **0.6** | 0.8 | 7.7 |
| fr3nostructure_near | | **2.4** | 2.7 | 7.5 |
| fr3nostructure_far | | **3.4** | X(ambiguity) | 18.3 |
| fr3structure_far | | **0.8** | **0.8** | 8.0 |
| fr3structur_near | | 2.5 | **1.6** | X |

Table 2. Quantitative Comparison on 10 TUM-RGBD sequences, comparing average keyframe position error, in cm.

| | | KF Trajectory RMSE (cm) / Per Frame Time (ms) | | |
|---|---|---|---|---|
| Sequence | # Frames | Ours | ORB | LSD |
| Desk_Circle | 1375 | 3.5 / 25.02 | **3.4** / 89.03 | 9.2 / 88.69 |
| Desk_Translation | 900 | **2.0** / 18.83 | **2.0** / 77.93 | 10.7 / 85.62 |
| Robot_Circle | 1200 | 2.7 / 20.14 | **1.8** / 92.29 | 6.8 / 109.4 |
| Forward_1 | 600 | **0.6** / 19.17 | 0.9 / 71.17 | 6.5 / 71.47 |
| Forward_2 | 700 | **1.2** / 19.45 | 1.4 / 74.08 | 4.0 / 87.55 |
| Forward_Backward | 2000 | 7.8 / 18.70 | **6.0** / 87.06 | 14.1 / 50.35 |
| Posters | 1750 | **2.0** / 24.17 | **2.0** / 83.71 | 5.2 / 86.74 |
| Desk_Circle_with Moving_Object | 2200 | **2.8** / 19.28 | 3.2 / 96.14 | FAIL / 39.52 |
| Boxes_Semi-Circle with_Moving_Object | 1700 | **1.3** / 19.20 | 3.5 / 87.90 | 12.0 / 58.01 |
| Lemniscate | 3200 | **3.4** / 21.96 | 12.1 / 95.51 | 52.8 / 86.69 |
| **Average** | | **2.7** / 20.59 | 3.6 / 85.38 | 13.5 / 76.40 |

Table 1. Quantitative Comparison on our 10 ground truth sequences, comparing average keyframe position error, in cm, and running times in ms.

SLAM and ORB-SLAM are cited from [32]. As shown in Table 2, we also achieve similar accuracy as ORB-SLAM. In the sequence fr3nostructure_far, ORB-SLAM refuses to start due to initialization ambiguity, but our method reconstructed this sequence successfully.

## 7. Conclusion

We present a novel monocular SLAM algorithm based on successes of recent global SfM algorithms. Our novel visual odometry step solves cameras and scene point depth together by a rank-1 factorization, which makes map initialization more robust. When combined with a global pose-graph optimization that separates the rotation, scale, and translation optimization, and computes each with robust incremental L1-optimization, we show that our approach can outperform recent state-of-the-art SLAM systems. Particularly, we have conducted experiments that indicate that our method outperforms ORB-SLAM and LSD-SLAM in terms of robustness to initialization, and G2O [27] and the EM based methods [30] in terms of dealing with false loops. The recovered camera position accuracy is generally on par with ORB-SLAM, and superior in difficult cases with forward motion and small parallax, which are common scenarios for many applications in robotics and automation.

**Acknowledgement** We thank Zhaopeng Cui for a lot of helps and discussions. This work is supported by the NSERC Discovery grant 611664, Discovery Acceleration Supplements 611663, and a research gift from Adobe.